\documentclass{llncs}
\usepackage{makeidx}  % allows for indexgeneration
\usepackage{graphicx}
\usepackage{svg}
\usepackage{amsfonts}
\usepackage{amsmath}
\usepackage{amssymb}
\usepackage{multirow}
\usepackage{booktabs,siunitx}
\usepackage{subcaption}
\usepackage{tabularx}
\usepackage{enumitem}
\usepackage{ulem}
\usepackage{cite}
\usepackage{hyperref}
\definecolor{instructioncolor}{rgb}{.0, .0, 0.0}
\definecolor{instcolor}{rgb}{.0, .0, .0}
\newcommand{\myung}[1]{\textcolor{instructioncolor}{#1}}
\newcommand{\myun}[1]{\textcolor{instcolor}{#1}}
\usepackage{etoolbox} % for "\patchcmd" macro
\usepackage{kotex}
\usepackage{xcolor}
\makeatletter
\patchcmd{\ps@headings}{\rlap{\thepage}}{}{}{}
\patchcmd{\ps@headings}{\llap{\thepage}}{}{}{}
\makeatother
\begin{document}
%
%\frontmatter          % for the preliminaries
%
%\pagestyle{headings}  % switches on printing of running heads
%\addtocmark{Hamiltonian Mechanics} % additional mark in the TOC

%\tableofcontents
%
\mainmatter              % start of the contributions
\title{TOSS: \myun{Real-time Tracking and Moving Object Segmentation for Static Scene Mapping}}
\titlerunning{TOSS}  % abbreviated title (for running head)
\author{Seoyeon Jang, Minho Oh, Byeongho Yu, I  Made Aswin Nahrendra, Seungjae Lee, Hyungtae Lim, \and Hyun Myung}
% \thanks{This research was supported by This work was supported in part by Korea Evaluation Institute of Industrial Technology (KEIT) funded by the Korea Government (MOTIE) under Grant No.20018216, Development of mobile intelligence SW for autonomous navigation of legged robots in dynamic and atypical environments for real application and by the KAIST Convergence Research Institute Operation Program. 
% The students are supported by BK21 FOUR.}

%
\authorrunning{Ivar Ekeland et al.} % abbreviated author list (for running head)
% \tocauthor{Ivar Ekeland, Roger Temam, Jeffrey Dean, David Grove,
% Craig Chambers, Kim B. Bruce, and Elisa Bertino}
%
\institute{School of Electrical Engineering, KAIST (Korea Advanced Institute of Science
and Technology), Daejeon 34141, Republic of Korea\label{inst:electronic}\\
\email{\{9uantum01, minho.oh, bhyu, anahrendra, sj98lee, shapelim, hmyung\}@kaist.ac.kr}\\  
\texttt{http://urobot.kaist.ac.kr}
}

\maketitle              % typeset the title of the contribution
\begin{abstract}
Safe navigation with simultaneous localization and mapping~(SLAM) for autonomous robots is crucial in challenging environments. To achieve \myung{this goal, detecting moving objects in the surroundings and building a static map are essential}. However, existing moving object segmentation methods have been developed separately for each field, making it challenging to perform real-time navigation and precise static map building simultaneously. 
In this paper, we propose an integrated real-time framework that combines online tracking-based moving object segmentation with static map building. For safe navigation, we introduce a computationally efficient hierarchical association cost matrix to enable real-time moving object segmentation. In the context of precise static mapping, we present a voting-based method, DS-Voting, designed to achieve accurate dynamic object removal and static object recovery by emphasizing their spatio-temporal differences.
We evaluate our proposed method quantitatively and qualitatively in the SemanticKITTI dataset and real-world challenging environments. The results demonstrate that dynamic objects can be clearly distinguished and incorporated into static map construction, even in stairs, steep hills, and dense vegetation.
\end{abstract}
%%%%%%%%%%%%%%%%%%%%%%%%%%%%%%%%%%%%%%%%%%%%%%%%%%%%%%%%%%%%%%%%%%%%%%%%%%%
\section{Introduction}
Autonomous navigation~\cite{voxblox} and simultaneous localization and mapping~(SLAM) \cite{shan2020lio},~\cite{Wei2021fastlio2} through mobile robots are crucial fields that allow us to acquire precise information about our desired environment. These two domains have a mutually dependent relationship. This is because efficient navigation planning enables the acquisition of better terrain information in the desired environment, building higher-quality static maps~\cite{beliveau1996autonomous}. These high-quality static maps, in turn, can assist in robot localization~\cite{zhang2014loam},~\cite{shan2018lego} and efficient path planning~\cite{voxblox},~\cite{hornung2013octomap}. 
% Recent research has expanded the scope of this field to include unstructured environments, which are inaccessible to humans~\cite{oh2022travel}, thanks to quadruped robots that offer greater maneuverability compared to wheeled robots.~\cite{step},\cite{nahrendra2023dreamwaq}.

Moving object segmentation~(MOS)~\cite{chen2021ral},~\cite{mersch2022ral} plays a crucial role in these two domains. From a navigation perspective, effective planning depends on accurately recognizing moving objects~\cite{TROT-Q},~\cite{schmid2023dynablox}, while from a mapping perspective, high-quality static maps can be generated when moving objects are effectively removed~\cite{lim2021erasor},~\cite{lim2023erasor2}. Despite the clear need for a MOS system capable of simultaneously addressing navigation and SLAM, previous research has traditionally divided its focus into MOS systems specialized for static mapping, localization, and navigation.

Static mapping-specialized MOS approaches~\cite{lim2021erasor},~\cite{kim2020remove} mainly operate offline so they are impractical for navigation purposes. On the other hand, localization-specialized~\cite{tian2022dl},~\cite{lin2023lio-segmot}, and navigation-specialized MOS methods~\cite{hornung2013octomap},~\cite{schmid2023dynablox} have limitations in terms of map quality. The former focuses more on improving localization accuracy, while the latter focuses more on building simplified maps for navigation purposes rather than providing a dense and clean map. Recently, frameworks~\cite{chen2021ral},~\cite{mersch2022ral} employing real-time deep neural networks for moving object segmentation and static mapping have been proposed. However, this approach requires significant computational resources.

In this paper, we present a novel MOS framework to overcome the limitations of previous studies and operate robustly, even in unstructured environments. Our contributions are as follows:
\begin{itemize}
\item[$\bullet$]{We propose TOSS, an online MOS system that integrates dynamic object tracking and real-time static mapping.}
\item[$\bullet$]{We have drastically reduced the tracking association time complexity from $\mathit{O(N^2)}$ to $\mathit{O(N)}$ by proposing an efficient hierarchical association cost matrix.}
% We linearized the quadratic tracking association time complexity
\item[$\bullet$]{Our novel approach, DS-Voting, which focuses on spatio-temporal disparities within tracked static and dynamic objects, substantially reduces false static and dynamic objects.}
\end{itemize}
%%%%%%%%%%%%%%%%%%%%%%%%%%%%%%%%%%%%%%%%%%%%%%%%%%%%%%%%%%%%%%%%%%%%%%%%%%%
\section{Related Works}
Moving object segmentation can be categorized into several areas depending on their primary objectives. Offline map cleaning methods~\cite{lim2021erasor},~\cite{lim2023erasor2},~\cite{kim2020remove} have been predominantly employed to create a precise static map. These methods involved constructing a pre-built map that includes dynamic and static objects, subsequently compared to each scan data. The map sections displaying significant discrepancies compared to the scan data were likely to represent dynamic objects. Consequently, a static map was generated by eliminating these dynamic objects. For example, Kim and Kim~\cite{kim2020remove} identified dynamic objects by analyzing the differences between two images derived from the map and scan data, which were organized into range images. On the other hand, Lim~\textit{et al.}~\cite{lim2021erasor} partitioned the map and scan data into grid areas and detected dynamic objects by comparing heights within each grid area.

On the contrary, navigation-specific methods identified moving objects to construct a map suitable for navigation. These methods included occupancy grids~\cite{elfes1989occupancy} or OctoMap~\cite{hornung2013octomap}. They divided the space into grid cells that can be either~\textit{occupied} or~\textit{free}. If specific points occupied a cell, its state was updated to \textit{occupied}; if not, it was updated to \textit{free}. Ultimately, consistently occupied cells were used to generate static maps. More recently, Schmid~\textit{et al.}~\cite{schmid2023dynablox} introduced a novel approach that divided the map based on truncated signed distance field (TSDF) grid cells, rather than occupancy grid cells, to construct a static map by determining whether the TSDF surface to which points belong has changed. However, map update methods had a significant computational cost as they required maintaining the state of all grid cells comprising the map in every frame.

Alternatively, methods aimed at improving the accuracy of robot localization used the states of moving objects in SLAM optimization to ensure more reliable localization even in highly dynamic environments~\cite{tian2022dl},~\cite{lin2023lio-segmot},\cite{qian2022rf}.

With the advancement of deep learning, recent studies have treated MOS as a task for deep learning models. Chen~\textit{et al.} trained LMNet using residual input images from 3D LiDAR range images in a continuous time frame~\cite{chen2021ral}. It had the advantage of operating quickly in real-time, but the quality of the residual images was significantly influenced by robot poses and data noise. Mersch~\textit{et al.} proposed a MOS method to learn motion features using a 4D sparse convolution network~\cite{mersch2022ral}.
\vspace{-0.1cm}
\section{Methodology}
\begin{figure}[!h]
    \centering
    % \includesvg[inkscapelatex=false, width=\textwidth]{01_fig_framework/framework_new.svg}
    \includegraphics[width=0.8\textwidth]{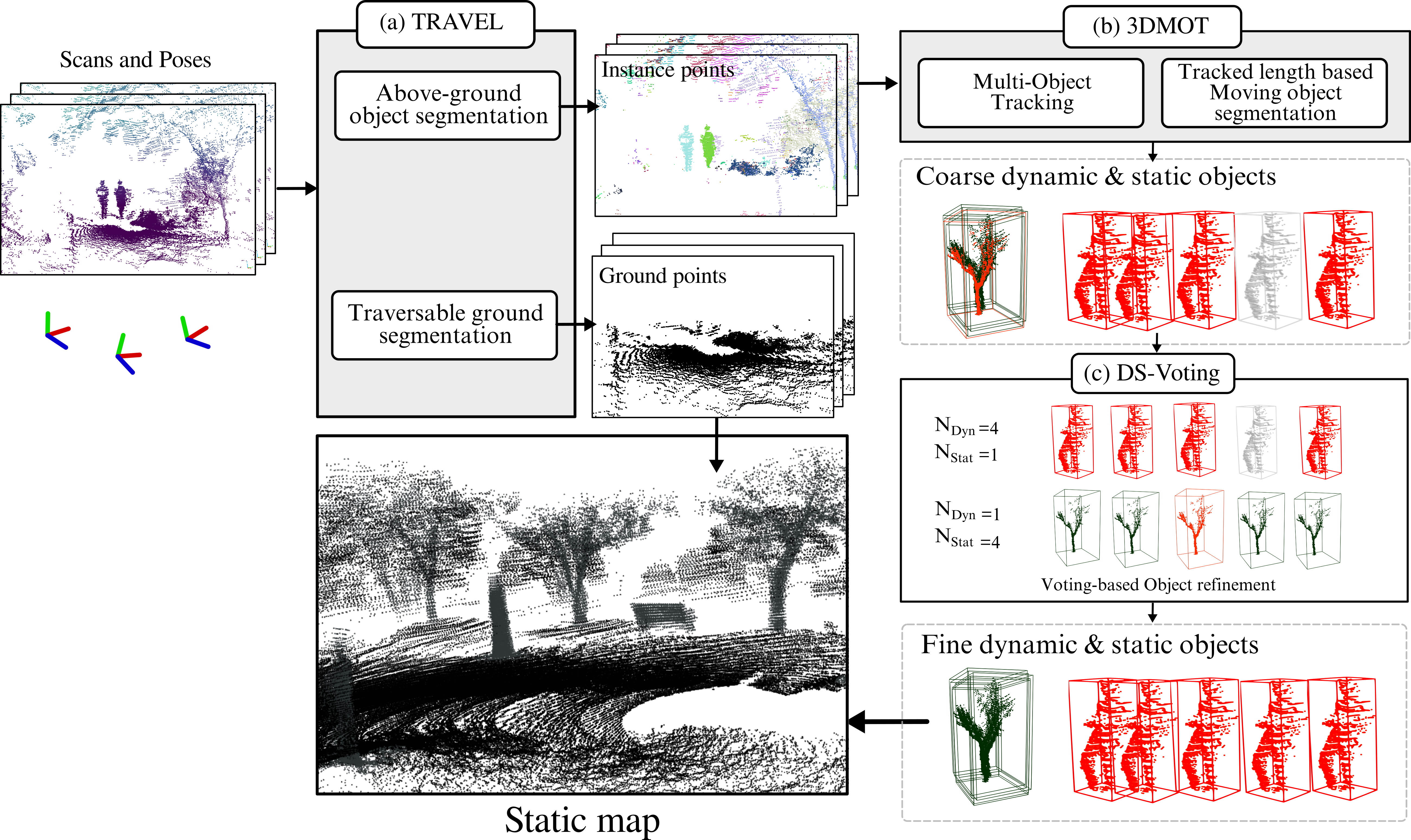}
    \caption{Overview of TOSS. TOSS is a multi-step process that involves (a) segmenting ground and instance points (see Sec.~\ref{sec:travel}), (b) real-time object tracking to classify them as coarse dynamic or static objects (see Sec.~\ref{sec:mot}), and finally (c) a voting-based refinement module for accurate dynamic object removal and static object recovery on the static map (see Sec.~\ref{sec:dsvote}).}
    \label{fig:travel_overview}
    % \vspace{-0.6cm}
\end{figure}
%%%%%%%%%%%%%%%%%%%%%%%%%%%%%%%%%%%%%%%%%%%%%%%%%%%%%%%%%%%%%%%%%%%%%%%%%%%
\subsection{Overview}\label{sec:overview}
In this chapter, we will describe our proposed approach for \myun{real-time \textbf{T}racking and moving object segmentation for~\textbf{S}tatic \textbf{S}cene mapping}, \textbf{TOSS}. TOSS tracks segmented objects, determines which elements are dynamic objects, and updates the static map using only static elements. To explain the process in more detail, it consists of three main steps.

First, the raw point cloud is input into (a) Traversable ground and instance segmentation module~\cite{oh2022travel}~(see Sec.~\ref{sec:travel}). Concurrently, we retrieve robot poses from existing LiDAR SLAM or odometry modules (see Sec.~\ref{sec:odom}). These instance points, originally in the local sensor coordinate system, are then transformed into a global coordinate system using the estimated poses. Subsequently, these transformed instance points are represented as bounding boxes.

Next, we apply (b) 3D multi-object tracking to these bounding boxes (see Sec.~\ref{sec:mot}). This process classifies them into dynamic and static box traces by comparing their maximum length and movement distances. However, it is important to note that occasional false negative points may arise due to tracking failures, and false positive points may result from incorrect motion detection.

To address these issues, we propose our novel module, (c) Dynamic-static voting (DS-Voting) (see Sec.~\ref{sec:dsvote}), to refine the initial dynamics and statics. Finally, we gather only the refined static points and ground points to construct the static map.
%%%%%%%%%%%%%%%%%%%%%%%%%%%%%%%%%%%%%%%%%%%%%%%%%%%%%%%%%%%%%%%%%%%%%%%%%%%
\subsection{Traversable Ground and Above-Ground Object Segmentation}\label{sec:travel}
To recognize and track objects located in various terrains, it is essential to separate ground points and cluster above-ground points into object units simultaneously. We employee a spherical projection-based instance segmentation algorithm proposed by Oh~\textit{et al.}~\cite{oh2022travel} First, it converts each point $p_i = (x, y, z)$ by mapping $\Pi : \mathbb{R}^3 \mapsto \mathbb{R}^2$ to spherical coordinates and finally to image coordinates~\cite{behley2018rss}, defined as follows:
\begin{equation}
 \label{eq:proj}
{u\choose v} = {{{1}\over{2}}[1-\text{arctan}(y,x)\pi^{-1}]~w\choose [1 - (\text{arcsin}(zr^{-1}) + f_{up})f^{-1}]~h}
\end{equation}

\noindent where $(u, v)$ are the image coordinates, $(h, w)$ are the height and width of the desired spherical image, $f = f_{up} + f_{down}$ is the vertical field-of-view of the LiDAR, and $r = ||p_i||_2$ is the range of each point. points structured with image coordinates can then be efficiently clustered through distance searches between neighboring points in both the horizontal and vertical directions. Finally, the clustered point cloud at time $t$ in the sensor frame $\mathcal{P}^{(t)}_\mathit{S}$ is divided as follows:
\begin{equation}
    % \mathcal{P^\mathit{(t)}_\mathit{S}} = \mathcal{G^\mathit{(t)}_\mathit{S}}  \cup  \mathcal{S^\mathit{(t)}_\mathit{S}}
    \mathcal{P}_{S}^{(t)} = \mathcal{G}_{S}^{(t)} \cup \mathcal{S}_{S}^{(t)},
\end{equation}
\begin{equation}
    \label{eq:insseg}
    \mathcal{S}_{S}^{(t)} = {\bigcup_{k=1...N}}~{S}_{S,k}^{(t)},
\end{equation}
where ground points are denoted as $\mathcal{G}^{(t)}_\mathit{S}$, segmented points as $\mathcal{S}^{(t)}_\mathit{S}$, and $N$ denote the number of instances estimated by TRAVEL.
%%%%%%%%%%%%%%%%%%%%%%%%%%%%%%%%%%%%%%%%%%%%%%%%%%%%%%%%%%%%%%%%%%%%%%%%%%%
\subsection{LiDAR Odometry and SLAM}\label{sec:odom}

In order to more accurately track instances within scan data at different times, it is necessary to transform each scan data on the egocentric perspective to the map perspective. However, noise or drift may occur when estimating the robot's pose. They can significantly weaken the performance of the map update method~\cite{elfes1989occupancy},~\cite{hornung2013octomap}, which requires accurate pose. On the other hand, our approach uses the estimated pose solely for the purpose of compensating for the robot's motion, so noise or drift does not matter significantly. As evidence, we show the results of using the estimated robot pose through the FAST-LIO~\cite{Wei2021fastlio2} and the SuMa~\cite{behley2018rss}.
%%%%%%%%%%%%%%%%%%%%%%%%%%%%%%%%%%%%%%%%%%%%%%%%%%%%%%%%%%%%%%%%%%%%%%%%%%%
\subsection{3D Multi-Object Tracking (3DMOT)}\label{sec:mot}

Next, we use Kalman filter-based 3D multi-object tracking (3DMOT)~\cite{weng2020ab3dmot} to distinguish between dynamic and static objects among the instances obtained by Eq.~\ref{eq:insseg}. 

All instance points, denoted as $\mathcal{S^\mathit{(t)}_\mathit{S}}$, are converted into global coordinates using the estimated pose $\mathbf{T}^\mathit{(t)}$ and represented as bounding boxes. The method for converting points into the bounding box follows the same procedure as {Auto-MOS} approach~\cite{chen2022automos}. That is,~$c$ is the center, $\theta$ is the heading angle, $l,w,$ and $h$ are the length, width, and height of the box, respectively.
\begin{equation}
% \mathcal{S^\mathit{(t)}_\mathit{G}} = {\bigcup_{k = 1...N}} {\mathbf{T}^\mathit{(t)} \mathcal{S^\mathit{(t)}_\mathit{S}}},
\mathcal{S^\mathit{(t)}_\mathit{G}} = {\bigcup_{k = 1...N}} \{{{\mathbf{T}^\mathit{(t)}~\mathbf{p}\mid\mathbf{p}\in\mathcal{S^\mathit{(t)}_\mathit{S,k}}}\}}, 
\end{equation}
\begin{equation}
{b^\mathit{(t)}_\mathit{G}} = \text{box}(\mathcal{S^\mathit{(t)}_\mathit{G}}) = [c_x, c_y, c_z, \theta, l, w, h]
\end{equation}

We associate the box detected at time $t$ with the trace boxes tracked from the previous time $t-1$ using a hierarchical cost matrix, which is a modification of {Auto-MOS}~\cite{chen2022automos}. The cost is a linear combination of the Euclidean distance cost between the center of the boxes $c_d$, the boxes' intersection-over-union cost $c_o$, and the bounding box volume cost $c_v$.  In {Auto-MOS}, the cost matrix $\mathcal{\mathbf{C} = \mathit{N^\mathit{(t)}} \times \mathit{N^\mathit{(t-1)}}}$~is calculated exhaustively for $\mathit{N^\mathit{(t)}}$ boxes and $\mathit{N^\mathit{(t-1)}}$ tracked boxes. 
\begin{equation}
\mathbf{C}_{i, j} = c_d + c_o + c_v,
\end{equation}
\begin{equation}
\mathit{c_d} = \|c_i - c_j\|_2,\end{equation} 
\begin{equation}
    \mathit{c_o} = 1 - IoU(b_i, b_j),
\end{equation}
\begin{equation}
    \mathit{c_v} = 1 - {\textrm{min}(\mathit{v_i, v_j}) \over \textrm{max}(\mathit{v_i, v_j})},
\end{equation}\\
here, $\mathit{c_i, c_j}$ represent the box centers, $\mathit{b_i, b_j}$ are the bounding boxes, and $\mathit{v_i, v_j}$ are the volume of the boxes. However, computing $N^\mathit{(t)} \times \mathit{N^\mathit{(t-1)}}$ cost matrices each time is a highly time-consuming task, it requires at least $\mathit{O(N^\mathit{(t)}\cdot{N^\mathit{(t-1)}})}$. 

To tackle this issue, we make the assumption that the center points of the same instance box at adjacent time frames will be very close in distance. Therefore, we select only the min($\mathit{k}$, $N^{(t)}$) nearest neighbor traces for the box at time $t$ as association candidates, where k denotes the $k$-th nearest bounding boxes in the $t$-th instances, which is a user-defined parameter. Subsequently, the cost matrix is computed only for these candidates, and the bounding box with the lowest cost is associated. This hierarchical calculation only needs to compare $N(t) \times \min(N^{(t-1)}, k)$ boxes, significantly reducing computational time to $\mathit{O(kN^{(t)})}$ while achieving the same performance as exhaustive matching. Detailed comparisons of computational time can be found in Table.~\ref{table:runtime} in Sec.~\ref{sec:result}.

%%%%%%%%%%%%%%%%%%%%%%%%%%%%%%%%%%%%%%%%%%%%%%%%%%%%%%%%%%%%%%%%%%%%%%%%%%%
\subsection{Dynamic-static Objects voting~(DS-Voting)}\label{sec:dsvote}
Determining whether an object is dynamic or static based solely on the length of a tracked object's trajectories can result in numerous misjudgments.

False negative points can arise due to tracking failures, and false positive points can result from inaccurate motion detection. To address this, we propose a novel approach named dynamic-static Voting (DS-Voting). This approach focuses on spatio-temporal differences for tracked dynamic and static objects.

The trajectories of dynamic objects exhibit strong associations primarily within neighboring regions of adjacent time frames, whereas the trajectories of static objects maintain consistent associations regardless of time. Formally, we define a bounding box $b^{(t)}$ and a flag function that specifies whether it is spatially close to any trace at time $t+k$ as follows:\\
\begin{equation}
    \mathit{f}(c^{(t)}, c_{\myung{\text{track}}}^{(t+k)}) = \begin{cases}1 & \parallel c^{(t)}  - c_{\myung{\text{track}}}^{(t+k)}\parallel_2~<~\tau\\0 & \text{otherwise}\end{cases}
\end{equation}
    \noindent where $c^{(t)}$ is the center point of the bounding box, $c_{\myung{\text{track}}}^{(t+k)}$ is the center point at time $t+k$ of any tracked object boxes, and $\tau$ is the threshold distance for two boxes to be associated with the same object. Based on this function, we calculate the number of times that the bounding box at time $t$ is dynamically counted as follows:
\begin{equation}
    N_{\myung{\text{dyn}}} = \sum_{\epsilon}~f(c^{(t)}, c_{\myung{\text{track}}}^{(t+\epsilon)})~\mid~\epsilon \in{[-\tau_{d}, \tau_{d}]}
\end{equation}
\begin{equation}
    N_{\myung{\text{stat}}} = \sum_{\epsilon}~f(c^{(t)}, c_{\myung{\text{track}}}^{(t+\epsilon)})~\mid~\epsilon \in{[0, -\tau_{s}],[\tau_{s}, N_{\myung{\text{frames}}}]}  
\end{equation}

\begin{figure}[t!]
    \centering
    % \includesvg[inkscapelatex=false, width=\textwidth]{08_DSVote/DSVote3.svg}
    \includegraphics[width=\textwidth]{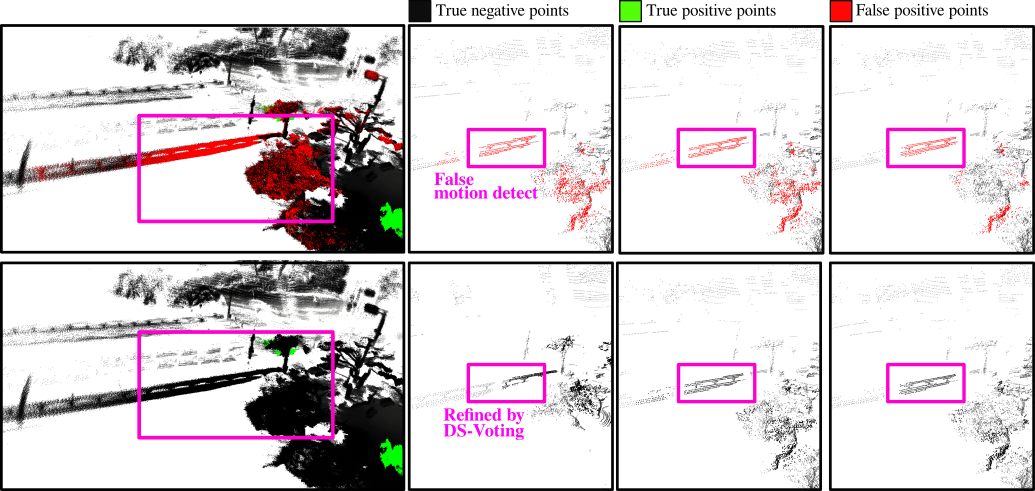}
    % \includesvg[inkscapelatex=false, width=\textwidth]{08_DSVote/SlideDT3.svg}
    \includegraphics[width=\textwidth]{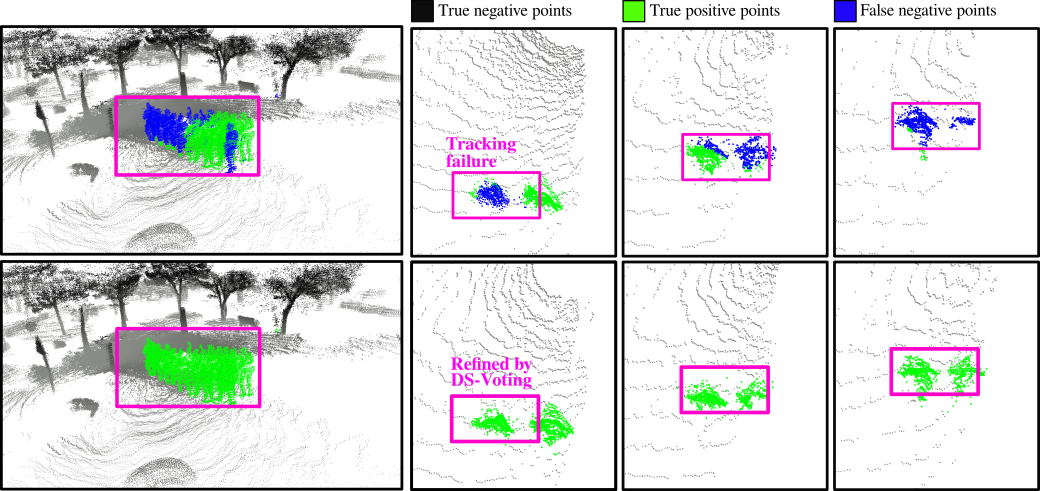}
    \vspace{-0.5cm}
    \caption{\textbf{\myung{Results comparison} for dynamic and static objects.} \myung{The first and third rows of the figure represent the results without using DS-Voting, while the second and fourth rows depict the results after using DS-Voting. false positive and false negative points are corrected by DS-Voting.}}
    \vspace{-0.4cm}
    \label{fig:dsvote-1}
\end{figure}

% \begin{figure}[h!]
%     \centering
%     \includesvg[inkscapelatex=false, width=\textwidth]{08_DSVote/SlideDT3.svg}
%     \caption{DS-Voting results for dynamic objects. Compared to the above result, DS-Voting can refine the false negative points marked in blue, as shown below.}
%     \vspace{-0.6cm}
%     \label{fig:dsvote-2}
% \end{figure}

\begin{equation}
    {\text{DS-Voting}(c^{(t)}) = \begin{cases}{c_{\myung{\text{stat}}}^{(t)}} & {\text{if}~N_{\myung{\text{dyn}}} < N_{\myung{\text{stat}}} }\\ c_{\myung{\text{dyn}}}^{(t)} & \text{otherwise}\end{cases}}
\end{equation}
If static objects exist in adjacent areas of dynamic traces recognized in time-continuous frames, it is highly likely that $N_{\myung{\text{dyn}}} > N_{\myung{\text{stat}}}$, so it is refined as dynamic objects. In addition, if traces with continuous motion are recognized even in the same spatial area of time-discontinuous frames, it is judged as static objects because it is highly likely that $N_{\myung{\text{dyn}}} < N_{\myung{\text{stat}}}$.

This approach improves the quality of the static map by reducing false negative points and false positive points, as shown in Fig.~\ref{fig:dsvote-1}.

%%%%%%%%%%%%%%%%%%%%%%%%%%%%%%%%%%%%%%%%%%5
\section{Experiments}\label{sec:experiment}
\subsection{Dataset}\label{sec:kitti} %%%%%%%%%%%%%%%%%%%%%%%%%%%%%%%%%%%%%%%%%%%%%%%%%%%%%%%%%

\textbf{SemanticKITTI dataset.} To evaluate the dynamic removal and static map generation performance of TOSS, we experimented with other static mapping based algorithms in SemanticKITTI benchmarks~\cite{behley2019semantickitti},~\cite{lim2021erasor}. The comparison algorithms are divided into map cleaning ~\cite{kim2020remove},~\cite{chen2022automos},~\cite{lim2021erasor},~\cite{lim2023erasor2}, and map updating~\cite{hornung2013octomap}, respectively. 
This benchmark is composed with five sequences with dynamic objects frequently appear: Seq. 00 (4390 - 4530), 01 (150 - 250), 02 (860 - 950), 05 (2350 - 2670), and 07 (630 - 820).

\begin{figure}[h!]
    \centering
    \includegraphics[width=\textwidth]{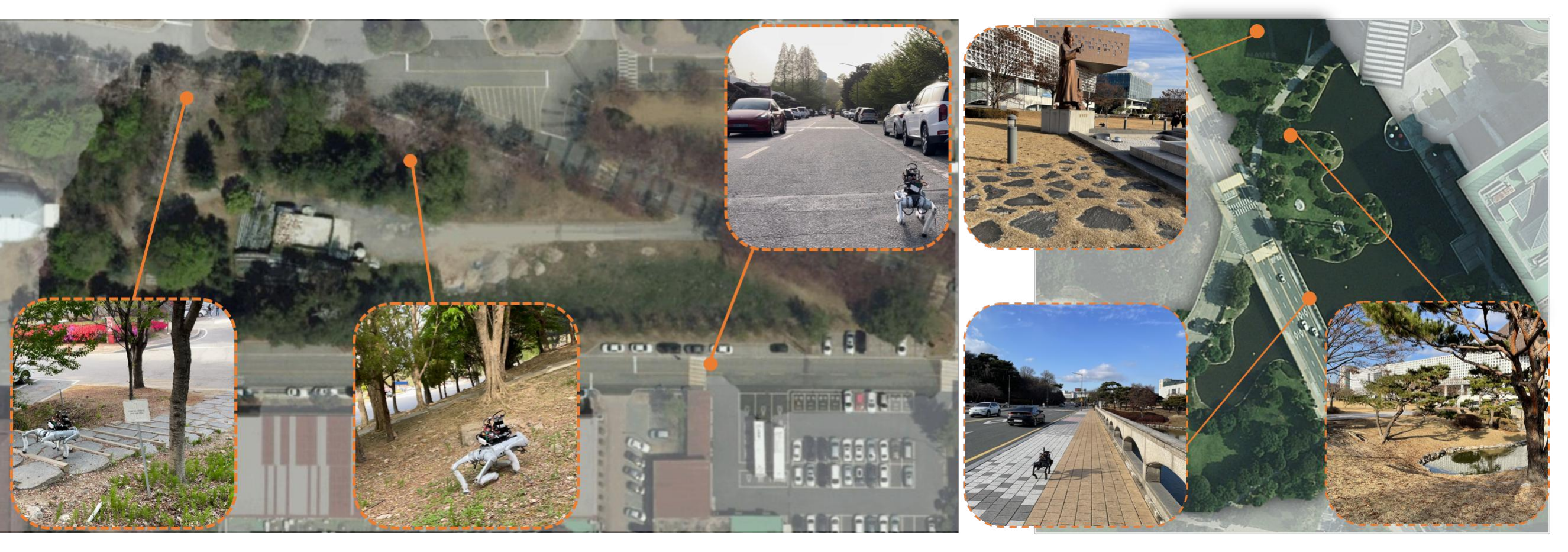}
    \caption{Our test environment for challenging environments in KAIST campus, \myung{R}epublic of Korea. Left test site ranges from flat road to stairways and steep hills in the wild, while right one contains various bumpy terrains.}
    \label{fig:environments}
\end{figure}

\begin{figure}[h!]
    \centering
    \includegraphics[width=\textwidth]{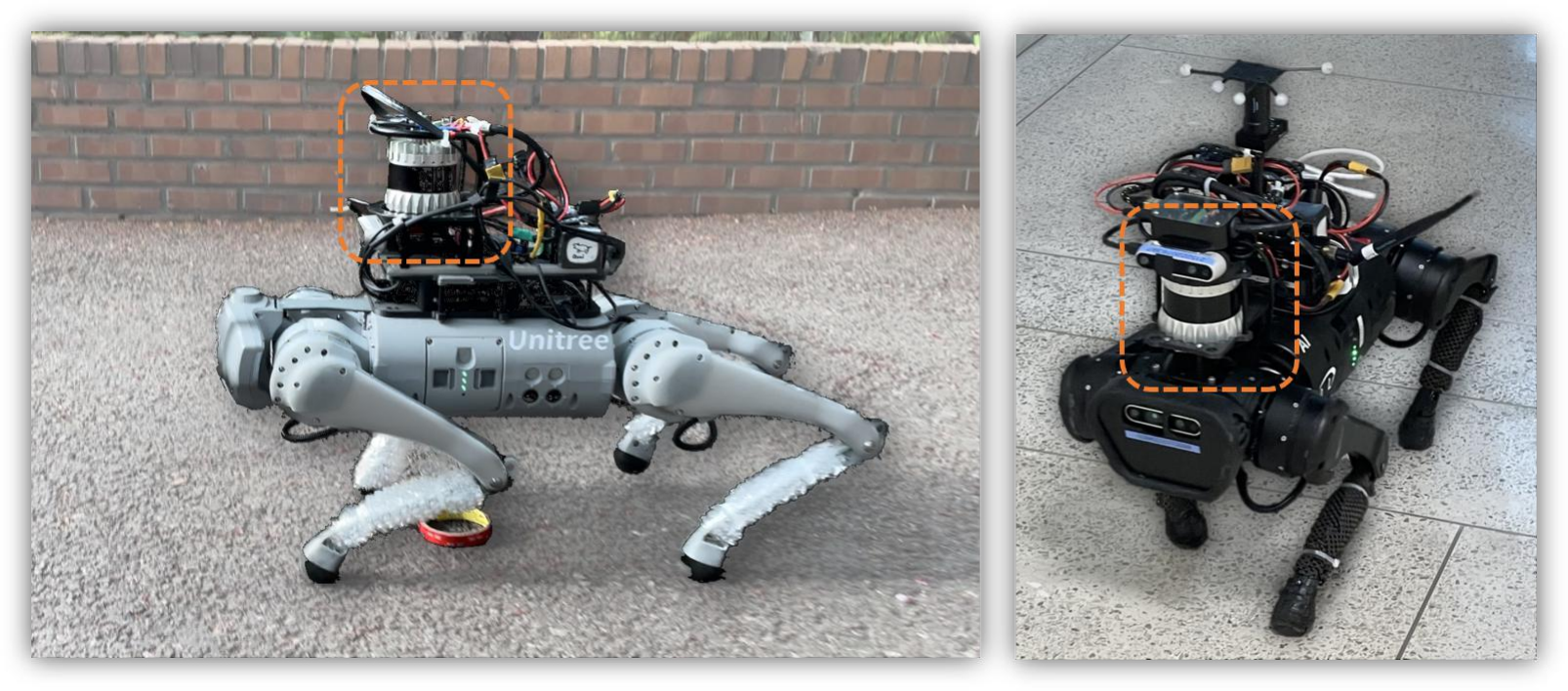}
    \caption{Our quadrupedal platforms, Go1 and A1 from Unitree Robotics, \myung{are} equipped with one range sensor~\myung{(Ouster OS0-128)} and one IMU sensor~\myung{(Xsens MTI-300)}.}
    \vspace{-0.6cm}
    \label{fig:platforms}
\end{figure}

\noindent \textbf{Real-world Unstructured Dataset.} To evaluate the performance of the proposed algorithm on the challenging situations such as climbing a hill, stairs, rough and bumpy terrains, our own rough terrain data introduces as shown in Fig~\ref{fig:environments}. 
The dataset was acquired via quadruped robots that walk on the challenging terrains. As shown in Fig.~\ref{fig:platforms}, these quadrupedal robot platform, Go1 and A1 from Unitree Robotics, are equipped with \myung{an} Ouster OS0-128 LiDAR sensor and \myung{Xsens MTI-300 IMU sensor}.
\vspace{-0.6cm}
 %%%%%%%%%%%%%%%%%%%%%%%%%%%%%%%%%%%%%%%%%%%%%%%%%%%%%%%%%%%%%%%
\subsection{Evaluation Metrics}
For quantitative static mapping performance evaluation, we utilize \textit{Preservation rate~(PR)},\textit{~Rejection rate~(RR)}, and \textit{F1 score~(PR)}, as proposed by Lim~\textit{et al.}~\cite{lim2021erasor}:
\begin{itemize}
    \item[$\bullet$] {$\mathrm{PR} = {\text{$\#$ of preserved static voxels} \over \text{num of total static voxels on the naively accumulated map}}$}
    \vspace{0.6mm}
    \item[$\bullet$] {$\mathrm{RR} = 1 - {\text{$\#$ of remaining dynamic voxels} \over \text{num of total dynamic voxels on the naively accumulated map}}$}
    \vspace{0.6mm}
    \item[$\bullet$] {$\mathrm{F_1} = \mathrm{2PR\cdot{RR} / {(PR + RR)}}$}.
\end{itemize}
% Those readers requiring comprehensive information regarding these metrics are encouraged to refer to~\cite{lim2021erasor}.

%%%%%%%%%%%%%%%%%%%%%%%%%%%%%%%%%%%%%%%%%%%%%%%%%%%%%%%%%%%%%%%%%%%%%%%%%%%%%%%%%%%%%%%%%%%%%%%%%%%%%
\section{Results}\label{sec:result}
As demonstrated in Table~\ref{table:semantic_kitti}, TOSS exhibits superior performance in terms of Preservation Rate (PR) compared to all map update and map cleaning algorithms. Particularly noteworthy is its performance in comparison to OctoMap~\cite{hornung2013octomap}, which employs the same map updating method. TOSS showcases significantly better PR performance, even in scenarios involving inaccurate pose estimations.\\
\vspace{-0.5cm}
\begin{table}[h!]
	\centering
	\caption{Quantitative evaluation against state-of-the-art methods on the static map benchmark using the SemanticKITTI dataset~(PR: Preservation Rate, RR: Rejection Rate).}
	{
        \scriptsize
	\begin{tabular}{l|c|lccc}
		\toprule 
            \midrule
		Seq. & Category & Method & \multicolumn{1}{c}{\begin{tabular}[c]{@{}c@{}} PR [\%] \end{tabular}}  & \multicolumn{1}{c}{\begin{tabular}[c]{@{}c@{}} RR [\%] \end{tabular}} & $\text{F}_{1}$ score \\ 
            \midrule
		%%%%%%%%% 00 %%%%%%%%%%
		\multirow{4}{*}{\texttt{00}}   
            & \multirow{2}{*}{Map Cleaning}
             & Removert - \texttt{RM3+RV1}~\cite{kim2020remove}                 & 86.829  & 90.617 &  0.887  \\ 
            && ERASOR~\cite{lim2021erasor}                                      & 93.980  & 97.081 &  0.955  \\
            \cline{2-6}
            & \multirow{2}{*}{Map Update}
            &\multirow{1}{*}{OctoMap - \texttt{0.2}~\cite{hornung2013octomap}} & 34.568  & \textbf{99.979} &  0.514            \\ 		
            && TOSS                                                             & \textbf{99.871} &  80.700 & \textbf{0.893}    \\
            \midrule
	% 	%%%%%%%%% 01 %%%%%%%%%%
		\multirow{4}{*}{\texttt{01}}
            & \multirow{2}{*}{Map Cleaning}
    		& Removert - \texttt{RM3+RV1}~\cite{kim2020remove}    &  95.815  & 57.077 &  0.715    \\ 
    		&& ERASOR~\cite{lim2021erasor}                        & 91.487 & 95.383  & 0.934      \\
            \cline{2-6}
            & \multirow{2}{*}{Map Update}
    		&\multirow{1}{*}{OctoMap - \texttt{0.2}~\cite{hornung2013octomap}} & 20.777  & \textbf{99.863} &  0.344          \\ 
            && TOSS                                                                 & \textbf{83.527}  & 93.733 &  \textbf{0.883} \\
            \midrule
		
	% 	%%%%%%%%% 02 %%%%%%%%%%
		\multirow{4}{*}{\texttt{02}}
            &  \multirow{2}{*}{Map Cleaning}
    	       & Removert - \texttt{RM3+RV1}~\cite{kim2020remove} & 83.293  & 88.371 &  0.858  \\ 
    	       && ERASOR~\cite{lim2021erasor}   & 87.731  & 97.008  & 0.921 \\
            \cline{2-6}
            &  \multirow{2}{*}{Map Update}
                & \multirow{1}{*}{OctoMap - \texttt{0.2}~\cite{hornung2013octomap}} & 23.746  & \textbf{99.792} &  0.384  \\ 
                && TOSS & \textbf{98.641}  & 98.783 & \textbf{0.987} \\
            \midrule
		
	% 	%%%%%%%%% 05 %%%%%%%%%%
		\multirow{4}{*}{\texttt{05}}
            & \multirow{2}{*}{Map Cleaning}
		   &\multirow{1}{*}{Removert - \texttt{RM3+RV1}~\cite{kim2020remove}} & 88.170  & 79.981 & 0.839  \\ 
		&& ERASOR~\cite{lim2021erasor} & 88.730  & 98.262  & 0.933  \\
            \cline{2-6}
            & \multirow{2}{*}{Map Update}
		&\multirow{1}{*}{OctoMap - \texttt{0.2}~\cite{hornung2013octomap}} & 33.904  & \textbf{99.882} &  0.506  \\ 
		&& TOSS & \textbf{99.208} & 63.714 & \textbf{0.776} \\
            \midrule
		
	% 	%%%%%%%%% 07 %%%%%%%%%%
            \multirow{4}{*}{\texttt{07}}
            & \multirow{2}{*}{Map Cleaning}
             & Removert - \texttt{RM3+RV1}~\cite{kim2020remove} & 82.038  & 95.504 &  0.883  \\ 
            && ERASOR~\cite{lim2021erasor} & 90.624  & 99.271  & 0.948  \\ 
            \cline{2-6}
            & \multirow{2}{*}{Map Update}
            &\multirow{1}{*}{OctoMap - \texttt{0.2}~\cite{hornung2013octomap}} & 38.183 & \textbf{99.565}  &  0.552  \\ 
            && TOSS & \textbf{99.732} & 95.913 & \textbf{0.978} \\ 
            \midrule  
            \bottomrule
	\end{tabular}
	}
	\label{table:semantic_kitti}
\end{table}
\begin{figure}[h!]
    \captionsetup{font=footnotesize}
    \centering
    \begin{subfigure}[t]{\columnwidth}
        \centering
        \includegraphics[width=0.9\textwidth]{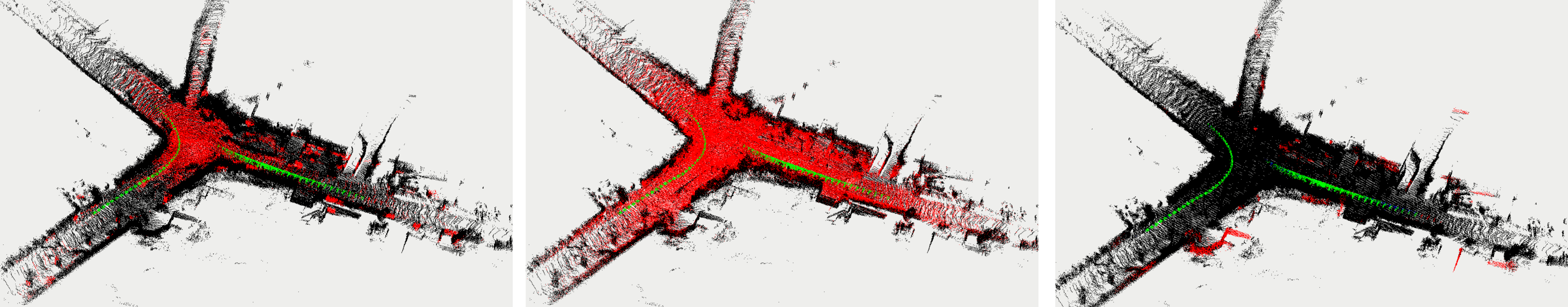}
    \end{subfigure}
    \centering
    \begin{subfigure}[t]{\columnwidth}
        \centering
        \includegraphics[width=0.9\textwidth]{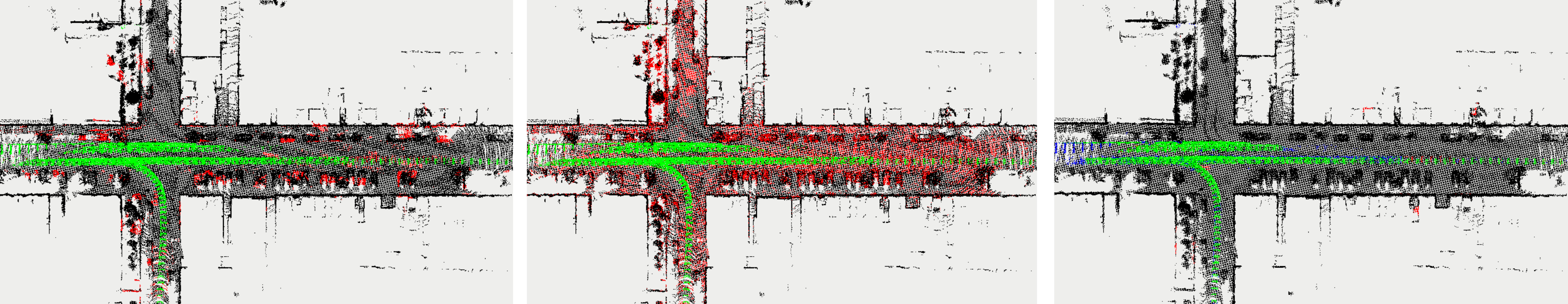}
    \end{subfigure}
    \vspace{-0.1cm}
    \caption{\textbf{Mapping result comparison {SemanticKITTI} sequence 02, 07.} Qualitative static map results ERASOR, OctoMap, and Ours. ERASOR is map cleaning method and OctoMap and Ours are map update method. Green points indicate true positive (TP), red points indicate false positive (FP), and blue indicates false negative (FN).}
    \label{fig:Semantickitti}
\end{figure}

However, our method has limitations. When object tracking encounters challenges related to instance over-segmentation or under-segmentation, dynamic traces may not be fully recognized. Consequently, our rejection rate performance may be relatively lower in certain sequences.
\begin{figure}[h!]
    \captionsetup{font=footnotesize}
    \centering
    \begin{subfigure}[t]{\columnwidth}
        \centering
        % \includesvg[inkscapelatex=false,width=\columnwidth]{07_mos/MOS5.svg}
        \includegraphics[width=\textwidth]{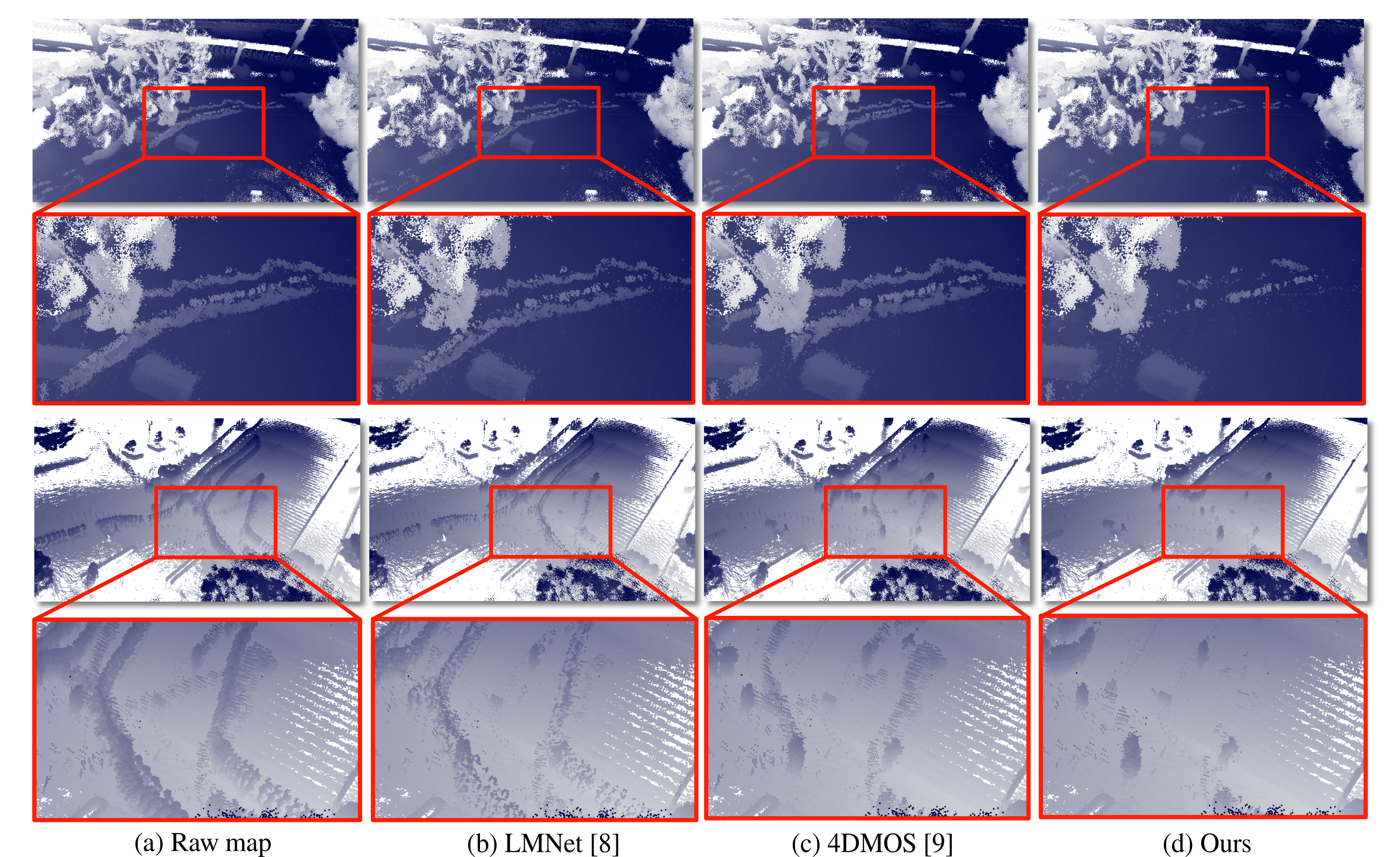}
    \end{subfigure}
    % \vspace{-0.1cm}
    % \begin{subfigure}[t]{\columnwidth}
    %     \centering
    %     \includesvg[inkscapelatex=false,width=\columnwidth]{07_mos/KI_mos2.svg}
    %     \caption{KAIST KI building long stairs}
    % \end{subfigure}
    \caption{\textbf{Comparison results with deep learning based MOS.} Our algorithm robustly tracks and erases dynamic objects even in challenging environments: KAIST duckpond and KAIST long stairs.}
    \label{fig:comparemos}
    \vspace{-0.3cm}
\end{figure}

We also conducted a comparison of our algorithm with real-time deep learning-based MOS methods.  LMNet~(Fig.~\ref{fig:comparemos}(b)) did not accurately remove dynamic points or removed too many static points. This behavior is primarily due to its susceptibility to robot pose variations when obtaining residual difference images.  On the other hand, in the case of 4DMOS~(Fig.~\ref{fig:comparemos}(c)), it was evident that dynamic objects could not be effectively removed in situations characterized by significant distortion caused by occlusion. 
In contrast, TOSS~(Fig.~\ref{fig:comparemos}(d)) utilizes Kalman filters for object tracking, enabling robust operation even in scenarios where occlusion occurs or robot pose estimations are inaccurate.
% \begin{figure}[h!]
%     \captionsetup{font=footnotesize}
%     \centering
%     \begin{subfigure}[t]{\columnwidth}
%         \centering
%         \includesvg[inkscapelatex=false, width=\columnwidth]{04_duckpond/duckpond_experiment.svg}
%         \caption{KAIST duckpond}
%     \end{subfigure}
%     \vspace{-0.1cm}
%     \begin{subfigure}[t]{\columnwidth}
%         \centering
%         \includesvg[inkscapelatex=false, width=\columnwidth]{05_KI/kI_experiment.svg}
%         \caption{KAIST KI building long stairs}
%     \end{subfigure}
%     \vspace{-0.1cm}
%     \begin{subfigure}[t]{\columnwidth}
%         \centering
%         \includesvg[inkscapelatex=false, width=\columnwidth]{03_north/north_experiment.svg}
%         \caption{KAIST north steep hills}
%     \end{subfigure}
%     \caption{\textbf{Mapping results on KAIST campus.} TOSS robustly tracks and erases dynamic objects even in steep hill and bush-covered environments. left one is point cloud map with dynamic traces, and right one is removed version of it.}
%     \label{fig:toss_kaist}
% \end{figure}

\begin{table}[h!]
\centering
\caption{Runtime comparison between Auto-MOS's exhaustive cost matrix calculation~\cite{chen2022automos} and our hierarchical approach. Detailed information about the experimental environment can be found in Fig.~\ref{fig:environments}.}
\vspace{1.6mm}
\begin{tabular}{cccc}
\hline
Methods  & Environments                 & \# Frames             & Runtime/frame {[}s{]} \\ \hline
Auto-MOS~\cite{chen2022automos} & \multirow{2}{*}{Duckpond}    & \multirow{2}{*}{501}  & 0.307               \\
Ours     &                              &                       & \textbf{0.029}               \\ \hline
Auto-MOS~\cite{chen2022automos} & \multirow{2}{*}{Long stairs} & \multirow{2}{*}{726}  & 0.223               \\
Ours     &                              &                       & \textbf{0.029}               \\ \hline
Auto-MOS~\cite{chen2022automos} & \multirow{2}{*}{Steep hills} & \multirow{2}{*}{4377} & 0.071               \\
Ours     &                              &                       & \textbf{0.014}               \\ \hline
\end{tabular}
\label{table:runtime}
\end{table}

\section{Conclusion}
In this paper, we have presented a real-time moving object segmentation and static map building system designed to operate robustly in unstructured environments. TOSS integrates instance segmentation, object tracking, and static map generation. As all these modules function in real-time, our system holds great potential for tasks demanding real-time autonomous navigation and the creation of static maps in challenging environments. Indeed, our real-world experimental results demonstrate that both dynamic and static objects can be robustly recognized and incorporated into static map creation, even in scenarios featuring long stairs, steep hills, and dense vegetation. 
% \myun{In the future works, we aim to enhance our segmentation and tracking module for identifying moving objects using deep learning approaches.}

\section{Acknowledgement}
\myun{This research was supported in part by the KAIST Convergence Research Institute Operation Program, and in part by Korea Evaluation Institute of Industrial Technology (KEIT) funded by the Korea Government (MOTIE) under Grant No.20018216, Development of mobile intelligence SW for autonomous navigation of legged robots in dynamic and atypical environments for real application.
The students are supported by BK21 FOUR.}

% and by Korea Evaluation Institute of Industrial Technology (KEIT) funded by the Korea Government (MOTIE) under Grant No.20018216, Development of mobile intelligence SW for autonomous navigation of legged robots in dynamic and atypical environments for real application. The students are supported by BK21 FOUR.
\newpage
\bibliographystyle{splncs}
\bibliography{./ref}

\begin{thebibliography}{10}

\bibitem{voxblox}
Oleynikova, H., Taylor, Z., Fehr, M., Siegwart, R., Nieto, J.:
\newblock Voxblox: Incremental {3D} euclidean signed distance fields for
  on-board {MAV} planning.
\newblock In: Proc. IEEE/RSJ International Conference on Intelligent Robots and
  Systems. (2017)  1366--1373

\bibitem{shan2020lio}
Shan, T., Englot, B., Meyers, D., Wang, W., Ratti, C., Daniela, R.:
\newblock {LIO-SAM}: Tightly-coupled {LiDAR} inertial odometry via smoothing
  and mapping.
\newblock In: Proc. IEEE/RSJ International Conference on Intelligent Robots and
  Systems, IEEE (2020)  5135--5142

\bibitem{Wei2021fastlio2}
Xu, W., Cai, Y., He, D., Lin, J., Zhang, F.:
\newblock {FAST-LIO2}: Fast direct {LiDAR}-inertial odometry.
\newblock IEEE Transactions on Robotics \textbf{38}(4) (2022)  2053--2073

\bibitem{beliveau1996autonomous}
Beliveau, Y.J., Fithian, J.E., Deisenroth, M.P.:
\newblock Autonomous vehicle navigation with real-time 3d laser based
  positioning for construction.
\newblock Automation in construction \textbf{5}(4) (1996)  261--272

\bibitem{zhang2014loam}
Zhang, J., Singh, S.:
\newblock {LOAM}: {LiDAR} odometry and mapping in real-time.
\newblock In: Proc. Robotics Science and Systems. Volume~2., Berkeley, CA
  (2014)  1--9

\bibitem{shan2018lego}
Shan, T., Englot, B.:
\newblock {LeGO-LOAM}: Lightweight and ground-optimized {LiDAR} odometry and
  mapping on variable terrain.
\newblock In: Proc. IEEE/RSJ International Conference on Intelligent Robots and
  Systems, IEEE (2018)  4758--4765

\bibitem{hornung2013octomap}
Hornung, A., Wurm, K.M., Bennewitz, M., Stachniss, C., Burgard, W.:
\newblock {OctoMap}: An efficient probabilistic {3D} mapping framework based on
  octrees.
\newblock Autonomous robots \textbf{34} (2013)  189--206

\bibitem{chen2021ral}
Chen, X., Li, S., Mersch, B., Wiesmann, L., Gall, J., Behley, J., Stachniss,
  C.:
\newblock Moving object segmentation in {3D LiDAR} data: A learning-based
  approach exploiting sequential data.
\newblock IEEE Robotics and Automation Letters \textbf{6} (2021)  6529--6536

\bibitem{mersch2022ral}
Mersch, B., Chen, X., Vizzo, I., Nunes, L., Behley, J., Stachniss, C.:
\newblock Receding moving object segmentation in {3D LiDAR} data using sparse
  {4D} convolutions.
\newblock IEEE Robotics and Automation Letters \textbf{7}(3) (2022)  7503--7510

\bibitem{TROT-Q}
Lee, E.M., Jeon, J., Myung, H.:
\newblock {TROT-Q}: Traversability and obstacle aware target tracking system
  for quadruped robots.
\newblock In: Proc. Asian Control Conference. (2022)  2480--2484

\bibitem{schmid2023dynablox}
Schmid, L., Andersson, O., Sulser, A., Pfreundschuh, P., Siegwart, R.:
\newblock Dynablox: Real-time detection of diverse dynamic objects in complex
  environments.
\newblock arXiv preprint arXiv:2304.10049 (2023)

\bibitem{lim2021erasor}
Lim, H., Hwang, S., Myung, H.:
\newblock {ERASOR}: Egocentric ratio of pseudo occupancy-based dynamic object
  removal for static {3D} point cloud map building.
\newblock IEEE Robotics and Automation Letters \textbf{6}(2) (2021)  2272--2279

\bibitem{lim2023erasor2}
Lim, H., Nunes, L., Mersch, B., Chen, X., Behley, J., Myung, H., Stachniss, C.:
\newblock {ERASOR2}: Instance-aware robust {3D} mapping of the static world in
  dynamic scenes.
\newblock In: Proc. Robotics Science and Systems. (2023)

\bibitem{kim2020remove}
Kim, G., Kim, A.:
\newblock Remove, then revert: Static point cloud map construction using
  multiresolution range images.
\newblock In: Proc. IEEE/RSJ International Conference on Intelligent Robots and
  Systems, IEEE (2020)  10758--10765

\bibitem{tian2022dl}
Tian, X., Zhu, Z., Zhao, J., Tian, G., Ye, C.:
\newblock {DL-SLOT}: Dynamic {LiDAR} {SLAM} and object tracking based on
  collaborative graph optimization.
\newblock arXiv preprint arXiv:2212.02077 (2022)

\bibitem{lin2023lio-segmot}
Lin, Y.K., Lin, W.C., Wang, C.C.:
\newblock Asynchronous state estimation of simultaneous ego-motion estimation
  and multiple object tracking for {LiDAR}-inertial odometry.
\newblock (2023)  1--7

\bibitem{elfes1989occupancy}
Elfes, A.:
\newblock Using occupancy grids for mobile robot perception and navigation.
\newblock Computer \textbf{22}(6) (1989)  46--57

\bibitem{qian2022rf}
Qian, C., Xiang, Z., Wu, Z., Sun, H.:
\newblock {RF-LIO}: Removal-first tightly-coupled {LiDAR} inertial odometry in
  high dynamic environments.
\newblock arXiv preprint arXiv:2206.09463 (2022)

\bibitem{oh2022travel}
Oh, M., Jung, E., Lim, H., Song, W., Hu, S., Lee, E.M., Park, J., Kim, J., Lee,
  J., Myung, H.:
\newblock {TRAVEL}: Traversable ground and above-ground object segmentation
  using graph representation of {3D} {LiDAR} scans.
\newblock IEEE Robotics and Automation Letters \textbf{7}(3) (2022)  7255--7262

\bibitem{behley2018rss}
Behley, J., Stachniss, C.:
\newblock Efficient surfel-based {SLAM} using {3D} laser range data in urban
  nvironments.
\newblock In: Proc. Robotics Science and Systems. (2018)

\bibitem{weng2020ab3dmot}
Weng, X., Wang, J., Held, D., Kitani, K.:
\newblock {AB3DMOT}: A baseline for {3D} multi-object tracking and new
  evaluation metrics.
\newblock arXiv preprint arXiv:2008.08063 (2020)

\bibitem{chen2022automos}
Chen, X., Mersch, B., Nunes, L., Marcuzzi, R., Vizzo, I., Behley, J.,
  Stachniss, C.:
\newblock Automatic labeling to generate training data for online {LiDAR}-based
  moving object segmentation.
\newblock IEEE Robotics and Automation Letters \textbf{7}(3) (2022)  6107--6114

\bibitem{behley2019semantickitti}
Behley, J., Garbade, M., Milioto, A., Quenzel, J., Behnke, S., Stachniss, C.,
  Gall, J.:
\newblock {SemanticKITTI}: A dataset for semantic scene understanding of
  {LiDAR} sequences.
\newblock In: Proc. IEEE/CVF International Conference on Computer Vision.
  (2019)

\end{thebibliography}

\end{document}